# Reinforcement Learning For Constraint Satisfaction Game Agents (15-Puzzle, Minesweeper, 2048, and Sudoku)

Anav Mehta

Cupertino High School

Cupertino, CA, USA

## Abstract

In recent years, reinforcement learning has seen interest because of deep Q-Learning, where the model is a deep convolutional neural network, trained with Q-learning variants. Deep Q-Learning has shown promising results in games such as Atari and AlphaGo. Instead of learning the entire Q-table, it learns an estimate of the Q function that determines a state's policy action. We use two reinforcement learning techniques, Q-Learning and deep Q-learning, to learn control policies of four constraint satisfaction games (15-Puzzle, Minesweeper, 2048, and Sudoku). Because of sparse rewards, Q-learning offers a better alternative than Policy Gradients with reward shaping. 15-Puzzle is a sliding puzzle that is a permutation problem and provides a challenge in addressing its large state space. Minesweeper and Sudoku involve partially observable game states and guessing. 2048 is also a sliding puzzle but allows for easier state representation (compared to 15-Puzzle) and uses interesting reward shaping to solve the game. These games offer unique insights into the potential and limits of reinforcement learning. The Q-agent is trained with no rules of the game, with only the reward corresponding to each state's action. Our unique contribution is in choosing the reward structure, state representation, and formulation of the deep neural network. For low shuffle 15-Puzzle, achieves a 100% win rate, the medium and high shuffle achieve about 43% and 22% win rates respectively. On a standard 16x16 Minesweeper board, both low and high-density boards achieve close to 45% win rate, whereas medium density boards have a low win rate of 15%. For 2048, the 1024 win rate was achieved with significant ease (100%) with high win rates for 2048, 4096, 8192 and 16384 as 40%, 0.05%, 0.01% and 0.004% , respectively. The easiest of Sudoku games had a win rate of 7%, while medium and hard games had 2.1% and 1.2% win rates, respectively. This paper explores the environment complexity and behavior of a subset of constraint games using reward structures which can get us closer to understanding how humans learn.

# 1. Introduction

Reinforcement learning (RL) [1] is one of the areas of machine learning - together with supervised and unsupervised learning - where agents interact (take actions) in an environment to maximize rewards. Supervised learning requires significant hand-labeled training data. On the other hand, RL algorithms learn from sparse, noisy, and delayed rewards (compared to the direct association between labels and targets in supervised learning). Secondly, supervised learning assumes the data to be independent, while in RL, sequences are highly correlated. We chose four games (Figure 1) with unique characteristics and RL strategies - state, reward, and deep Q-Learning (DQL) frameworks. There has been little work done in RL and constraint satisfaction games.

Figure 1: Games (Source: Wikipedia images)

| 15-Puzzle(a) | Minesweeper(b) |
|---|---|
| | |

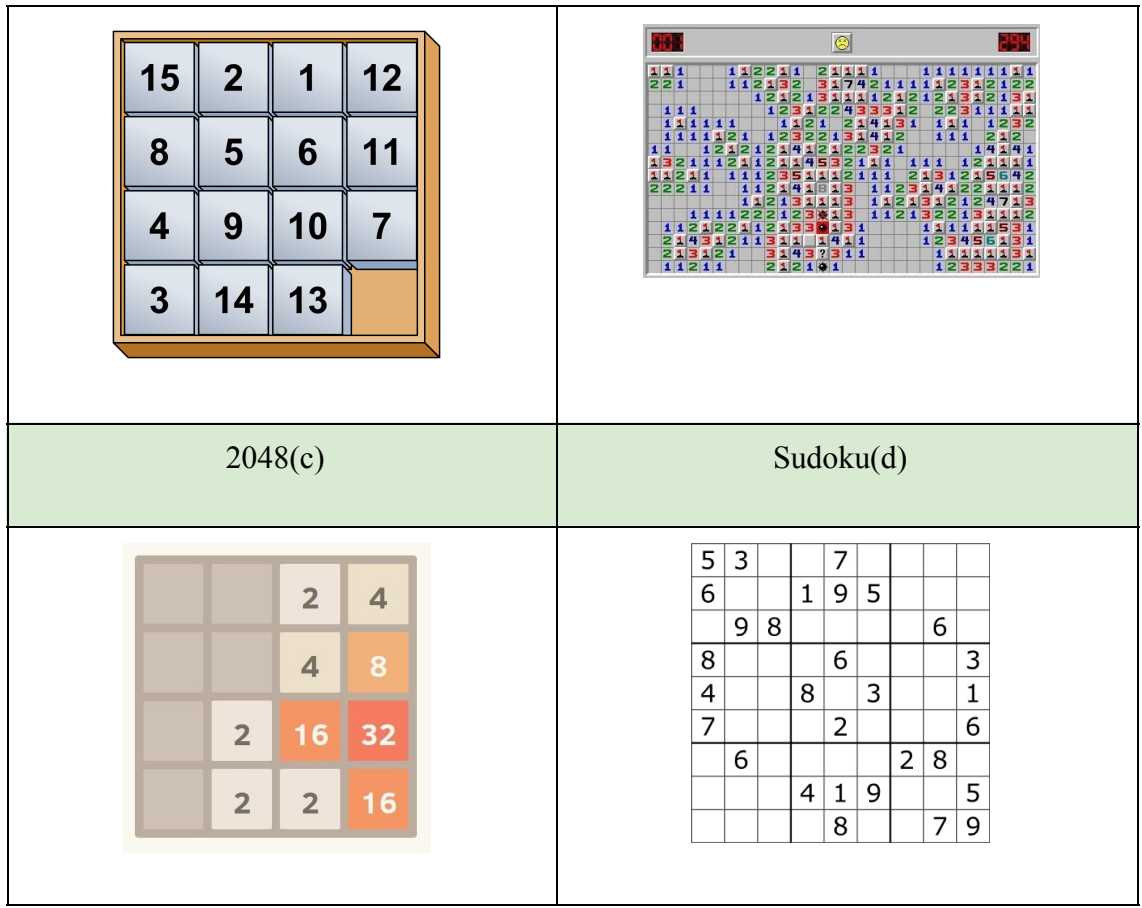

**15-Puzzle** [4] (Figure 1(a)) is a sliding puzzle that has numbered square tiles (1-15) in random order with one empty tile. The aim is to reorder the tiles by sliding the tiles. The challenge for RL lies in its representation of states.

**Minesweeper** [5] (Figure 1(b)) is a single-player puzzle. The goal is to clear a minefield with hidden mines without exploding none of them, with help from clues of the number of neighboring mines. RL's challenge is because its state and action space are large, it involves guessing, the next state can be indeterminate, and constraints spread over many cells.

**2048** [6] (Figure 1(c)) is a single-player sliding 4x4 grid puzzle game, where each cell is a power of two. Four sliding actions move cells in that direction. Two adjacent cells with the same value along that direction merge to a single cell with double the amount which continues until reaching 2048 or more. On every action, it inserts a two in a vacant cell. 2048 offers a straightforward RL state and action formulation, and merging happens in their channels.

**Sudoku** [7] (Figure 1(d)) is a puzzle where the objective is to fill a partial 9×9 grid with digits in each column, row, and in each of the nine 3×3 sub-grids contain numbers from 1 to 9. A Sudoku puzzle has a single solution, which makes the puzzle hard for RL since it is an exact cover problem and one and only one set of actions are applicable.

## 2. Previous Work

The most prominent work in Deep Q-Learning has been done by Google DeepMind [1], where an agent was successful in playing seven of the Atari 2600 games and surpassing human expert-level in three games. The agent could be trained despite high dimensional inputs and no information about the game.

**15-Puzzle:** One of the few papers in RL attempted 15-Puzzle using local value iteration [8].

**Minesweeper:** Previous attempts [9, 10] have used reinforcement learning with Q-Learning in both the classical and neural network versions. Neither of these approaches achieves more than a 5% win-rate on 6×6 boards with varying numbers of mines.

**2048:** A successful attempt at learning the game 2048 by [2] using Expectimax(a game theory algorithm) optimization and achieved tiles of 8192 on almost all games. Their model performs maximization over all the moves, followed by expectation over all possible tiles. Silver et al. [3] used the Monte-Carlo Tree search strategy to encourage optimal decision during the exploratory phase, but this turned out to be very expensive and did not produce successful results. Szubert and Jaśkowski [19] achieved a winning rate of 97% and used the N-tuple network and Temporal

Difference (TD). Wu et al.[20] adapted the agent to several stages of the game, using Multi-Stage Temporal Difference (MSTD).

**Sudoku** There is no RL work done in Sudoku, but the most promising relevant work is Recurrent Relational Networks [11]. It involves learning to solve tasks that require a chain of interdependent steps, like solving puzzles where the smaller elements of a mutually constrained solution. This method can solve Sudoku puzzles from supervised training data; however, this is still using supervised learning and not reinforcement learning.

## 3. Methods and Tools

### 3.1 Overview Of Reinforcement Learning Methods

We present the overview of RL methods to be used in our games. In RL, an agent learns from experience by taking actions and observing rewards to find an optimal policy. Due to sparse rewards, it is not possible to directly understand the policy network through policy gradients. Hence, we focused our efforts on q-learning methods. In three out of four games, we used a function approximation of the q-table as represented by the deep Q-network (DQN).

### 3.1.1 Markov Decision Process

The Markov Decision Process(MDP) [1] is used to model decision-making in stochastic environments. It is a 5-tuple ($S, A, P_a, R_a, \gamma$). $S$ and $A$ are a finite set of states and actions. $P_a(s, s')$ is the probability that action $a$ in state $s$ in time $t$ will result in state $s'$ in time $t+1$ (Markov property). $R_a(s, s')$ is the immediate reward received after a transition from state s to state $s'$ by taking action $a$ and $\gamma \in [0, 1]$ is the discount factor between future and present rewards. The agent takes an action $a_t$ depending only on the transition probability $P_a(s, s')$, observing an immediate reward $r_t$ and changing the state from $s_t$ to $s_{t+1}$. (Figure 2). $R_t$ is the sum of future rewards, where the next reward decreases with discounted factor $\gamma$.

$$R_t = r_t + \gamma r_{t+1} + \gamma^2 r_{t+2} + \dots + \gamma^{t+n} r_{t+n}$$

**Figure 2: Agent with MDP[1]**

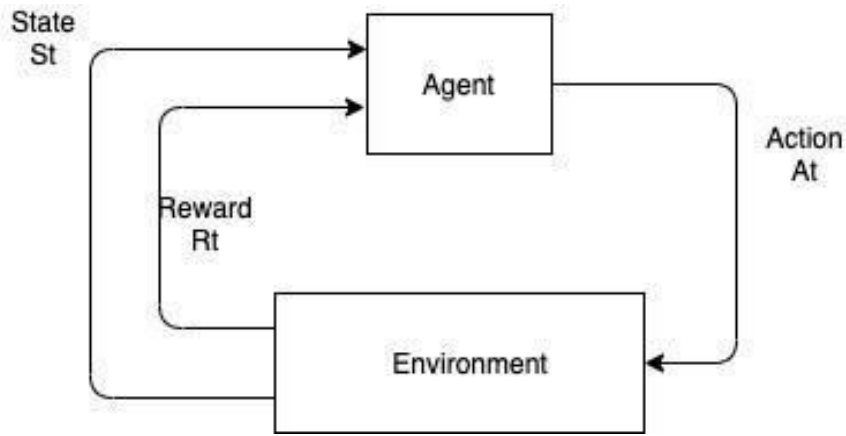

### 3.1.2 Q-Learning

An agent performs actions that maximize the discounted future reward $R_t = r_t + \gamma R_{t+1}$. The optimal action-value function $Q^\star(a, s)$ is the maximum expected reward by following a policy $\pi$. $\pi^\star$ is then the optimal policy [1] (Figure 3).

$$Q^\star(s, a) = \max_\pi E[R_t | s_t = s, a_t = a, \pi]$$

$$\pi^\star(s_t) = \arg\max_{a_t} Q^\star(s_t, a_t)$$

The optimal Q-function then follows the Bellman equation

$$Q^\star(s_t, a_t) = E[r_t + \gamma \max_{a_{j+t}} Q^\star(s_{t+1}, a_{t+1}) | s_t, a_t]$$

A model (of transition probabilities) is not available but is discovered via trial and error to derive the optimal policy (**off policy**). For the discovery process, we do not store any explicit policy, only a value function. The policy is here implicit and can be derived directly from the value function by picking the action with the best value.

**Figure 3: Q-Learning [1]**

Algorithm parameters: step size $\alpha$, $\epsilon$-greedy, discount factor $\gamma$ all $\in (0, 1)$
Algorithm variables: Q - Q-table, $s \in S$ - State Representation, $a \in A$ - Action Space, R - Reward
Initialize $Q(s, a)$, for all $s \in S^+$, $a \in A(s)$, arbitrarily except that Q(terminal, ·)=0

**Loop for each episode:**
   Initialize S
   Loop for each step of the episode:
   Choose A from S using policy derived from Q (e.g., $\epsilon$-greedy)
   Take action A, observe R, S'.
   $Q(S, A) \leftarrow Q(S, A) + \alpha[R + \gamma \max_a Q(S', a) - Q(S, A)]$
   $S \leftarrow S'$
**until S is terminal**

---

### 3.1.3 Deep Q Network

In practice, we use a DQN with parameters $\theta$ as a function approximator of the action-value function, $Q_\theta(s, a) = Q^\star(s, a)$ [4]. A Q-network can be trained by adjusting the parameter $\theta_i$ at iteration $i$ to reduce the mean-squared error in the Bellman equation. The loss L is

$$L_i(\theta_i) = [r_t + \gamma \max_{a_{t+1}} Q_{\theta_i}(s_{t+1}, a_{t+1}) - Q_{\theta_i}(s_t, a_t)]^2$$

DQN is used in continuous policy spaces, and the action space can be continuous or discrete. Moreover, since DQN is based on Q-learning, it is model-free and updates its policy after episodic tasks.

### 3.1.4 Experience Replay

A primary DQN agent learns from every experience and does not process it further. However, experiences are strongly correlated, and the agent does not experience sufficient variance. Hence, it is necessary to generate random experiences to jump-start the training process by storing experiences in a buffer pool. After the initial experience, we start training, which is more 'episodic' in nature. The agent's experiences at each time-step, $e_t = (s_t, a_t, r_t, s_{t+1})$ in a data-set $D = e_1, ..., e_N$, pooled over episodes into a replay memory [12, 13].

**Table 2: Overview of Algorithms Considered**

| Algorithm | Model | Policy | Action Space | Policy | Operator |
|---|---|---|---|---|---|
| Q-learning | N | Off | Discrete | Discrete | Q-value |
| DQN | N | Off | Discrete/Continuous | Continuous | Q-value |

## 3.2 Methods Used In Our Games

This section will identify the RL methods used in our chosen games, 15-Puzzle, Minesweeper, 2048, and Sudoku.

### 3.2.1 15-Puzzle

15-Puzzle is a simple sliding permutation puzzle with a huge state space ($(16)! = 2^{1013}$ different states). Optimal solutions can take up to 80 moves. We use some simple insights on how humans would solve this problem [15] to tackle the state space. The algorithm chosen is Q-value iteration, since the game state is visible throughout and does not require any approximation. The game difficulty is classified as the number of shuffles (i.e., i.e., the sum of the distances between the piece locations from their final position). The difficulty is classified as **easy** (shuffles < 10), **medium** (10 <= shuffles < 30) or **hard** (shuffles >= 30). Pizlo and Li [15] showed that humans learn the state of the 15-puzzle by trying to solve the puzzle tile by tile locally. While this approach does not provide optimal solutions, it significantly reduces the complexity of the problem. The basic principle is to hierarchically divide the global problem into small local problems by moving each tile to its correct position sequentially. We present an RL approach using local state space information, similar in [16, 17]. Thus, when learning a single tile policy, we consider only a limited number of possible puzzle configurations. The complete solution is arrived at by sequential application of the local solutions. A local sub-problem can be defined by a set $G = \{i_1, \ldots, i_k\} \subseteq \{1 \ldots 15\}$ of $k$ tiles that has to be moved to distinct positions $i_1 \ldots i_k$ without moving tiles $R = \{j_1, \ldots, j_l\}$. The G, R-subproblem is to bring tiles G to its correct position without moving tiles R. Here, $R \cap G = \varnothing$ holds. The state is a concatenated string of grid values with a ','. It starts sequentially addressing each piece, expanding the state space gradually when there is an episode loss when the pieces cannot be moved to their destination before a fixed number of moves. The game has mini-episodes, which place each number and previous numbers in its position (Figure 4). A simple sliding action tuple of up, right, down, and left moves the empty square. For the original state, space rewards are **no_progress**(-1), where the state remains the same, **progress** (+1) for every correct location improvement, **win** (+15) for reaching all the right locations, and **loss** (-15) when pieces cannot move to their final positions in 256 ($16^2$) moves. The modified state space has only one set of rewards where the mini-episode moves just a subset of tiles to the correct location. With this modified state, the rewards are **win** (+1), the tile and previous number tiles move to the final positions within tile square moves) and **loss** (-1), the tile and previous number tiles move to the final positions within tile square moves). E.g., for mini-episode 4, the goal is to put all 1-4 pieces in their right places within $4^2$ moves). The hyperparameters are **learning_rate** $\alpha$ (set to 0.5), **discount_factor** $\gamma$ (set to 0.7), **epsilon-greedy** $\epsilon$ (set to 0.1), and **episodes** (set to 1,000,000).

**Figure 4: Modified Q-Learning for Adaptive State for 15-Puzzle**

---

Variables $G_i$ are the tiles to be moved to their location, $R_i$ are fixed tiles.
s ← start state
**for** i ← 1, . . .15 **do**
  $G_i$ ← i, $R_i$ ← 1, ..., i − 1 //Successively move tiles to correct positions
  $Q^{Gi,Ri}$, $\pi^{Gi,Ri}$ ← q iteration($S^{Gi,Ri}$)
  **while** $Q^{Gi,Ri}(s) = 0$ **do** // holds iff, it is not possible to solve $G_i$, $R_i$ for s
    $G_i = G_i \cup \max(R_i)$ // move tile with the highest number from set $R_i$ to $G_i$
    $R_i = R_i \setminus \max(R_i)$
    $Q^{Gi,Ri}$, $\pi^{Gi,Ri}$ ← q iteration($S^{Gi,Ri}$)
  **end while**
  //Now solve the part of the puzzle corresponding to $G_i$, $R_i$
  **while** $i \in G_i$: $s(i) \neq i$ **do** // while at least one tile of $G_i$ is not at its goal position
    $a_{best}$ ← $\pi^{Gi,Ri}(s)$ // Policy $\pi$ returns the action to solve $G_i$, $R_i$
    $s \leftarrow f(s, a_{best})$ // Apply sliding action
  **end while**
**end for**

### 3.2.2 Minesweeper

Minesweeper's constraints can be modeled as an image representation, and hence we chose Deep Q-Learning as our algorithm. The game difficulty was classified using the percentage of mine density (i.e., #mines/NxM in a NxM minefield, in our case $N = 16$ and $M = 16$). The classifications were **low** (density < 35%), **medium** (35% <= density < 70%) and **high** (density >= 70%). The following were the state representations: a) **one_hot**, a 1-hot encoded $N \times M \times 10$ matrices (the first eight channels (1-8), the 9th channel is one if empty, and the 10th channel is if one field is unknown), b) **condensed**, a $N \times M \times 2$ matrix (the first channel is the integers (1-8), the second channel is one, if the field is unknown, otherwise 0) and c) **image** - N×M×1 matrix (one channel with 'color' of a field (0-9)). The **condensed** representation was much better since free minefields are part of the constraints. The action is pressing location $(i, j)$ on the NxM grid, and its value is $(M-1) * i + j$ ranging from 0 to $(N-1) * (M-1)$. The following was the reward structure: **win** (*NxM*) when the state of the entire minefield is determined, **loss** (-*NxM*) when the agent picks a cell that is a mine and loses, **progress** (1.0) when the agent picks a cell that is not a mine, **no_progress** (-0.5) when the agent picks an already known cell and finally **once** (-0.5) when a cell is picked and has 0 revealed neighbors (penalizes random guesses). Our deep Q-Network has three consecutive filters of size 4x4x16, 3x3x32, 2x2x64 are applied to input 16x16x2, concatenated to 16384, and then to the action space of 16x16 (or 256) (Figure 5(a)). The hyperparameters are **learning_rate** $\alpha$ (set initially to 0.001 and decreases by 1/10 and by 1/50), **discount_factor** $\gamma$ (set to 0.0, and 0.99 to give insight into agents' ability to use future rewards - 0.0 was eventually chosen), **epsilon_greedy** $\epsilon$ (annealed from 1 to 0.1 over 250,000 iterations), **episodes** (set to 1,000,000) and **buffer_Q**, the experience buffer replay size (set to 100,000).

### 3.2.3 2048

For 2048, the state space is enormous. However, for each power of 2 channels, the state can be treated as an image independently, and DQL suits very well. The following state representations were considered: **one_hot,** where the state is a 1-hot encoded 4x4x16 matrix, and the value is a power of 2 from $2^0$ (empty cell) to $2^{15}$ with the total state space size $\{0, 1\}^{16}$ and **image,** where one channel holds a 'color' [0..15] of the log of value of the cell. We used the **one_hot** representation for our experiments because the processing happens in each of the individual spaces representing the power of 2, i.e., two neighboring values of $2^{k-1}$ are combined to form $2^k$. A 4-tuple (up, right, down, left) was simple enough to characterize the actions. Four features were used in the reward evaluation function ($R$): $x_0$ the maximum number in the grid, $x_1$, the number of free tiles, $x_2$, the number of adjacent cells with the same number and $x_3$, the sum of neighboring cells and their corresponding weights ($w_0$, $w_1$, $w_2$, $w_3$). Weights were ($w_0$, $w_1$, $w_2$, $w_3$ = (4, 3, 1, 1), with the most weight to maximum tile and free tiles.

$$R = w_0 \times x_0 + w_1 \times x_1 + w_2 \times x_2 + w_3 \times x_3$$

There is no negative reward if the player succeeds in winning (defined as achieving 1024). Our Deep Q-Network has three hidden layers, with two-level CNNs and a dense layer. The first level applies filters (1, 2) and (2, 1) to detect consecutive vertical and horizontal segments in different channels and increase the depth to 128. The second level has four CNN applying filters (1, 2) and (2, 1). Each stage has a ReLU, and the outputs of the final CNNs are concatenated into a dense network of size 6144 and then to 4. (Figure 5(b)). The hyperparameters are **learning_rate** $\alpha$ (set initially to 0.0005, and decayed exponentially every step, using the staircase and a decay rate of 0.9), **discount_factor** $\gamma$ (set to 0.9), **epsilon_greedy** $\epsilon$ (set to 0.9 and decreased every 2500 iterations by 99.5%), **episodes** (set to 200000), and finally **buffer_Q,** the experience buffer replay size (set to 100,000).

### 3.2.4 Sudoku

We believed that the constraints could be presented as an image, and hence we chose the algorithm used for Sudoku to be Deep Q-Learning. The number of guesses required determines the level of difficulty, **easy** (guesses <3), **medium** (3 <= guesses < 6), and **hard** (6 <= guesses). The state representation is the entire 9x9x10 space corresponding to the 9x9 matrix and 0-9 numbers. The action is the location value (9x9) and the entry value (1-10). The size of the action space is 9x9x10 or 810. The reward structure has four components: **no_progress** (-1), when a picked entry was filled (in the initial setup or by the agent), **progress** (+1), if an entry satisfies the Sudoku constraints, **partial_loss** (-2), if an entry does not satisfy the Sudoku constraints and finally **loss** (-20) if the number of moves per episode passes a threshold. This threshold is programmable and was chosen to be 81*2 to allow for corrections in multiple guesses. The DQN is a fully connected 810 state value that connects an 810 action output with a **matmul** followed by a ReLU (Figure 5(c)). The hyperparameters are l**earning_rate** $\alpha$ (starts at 0.001 and decreases by 1/10 and by 1/50), **discount_factor** $\gamma$ (set to 0.0), **epsilon_greedy** $\epsilon$ (annealed from 1 to 0.1 over 300,000 iterations), **episodes** (set to 300,000) and finally **buffer_Q** the experience buffer replay size (set to 100,000).

### 3.3 Tools

We developed the reinforcement learning environment of these games in Google Colab Pro using Python and Tensorflow using K80 Nvidia GPUs. For Minesweeper, 2048, and Sudoku, the modified OpenAI Gym environment was used.

# Figure 5: Deep Q-Networks Used

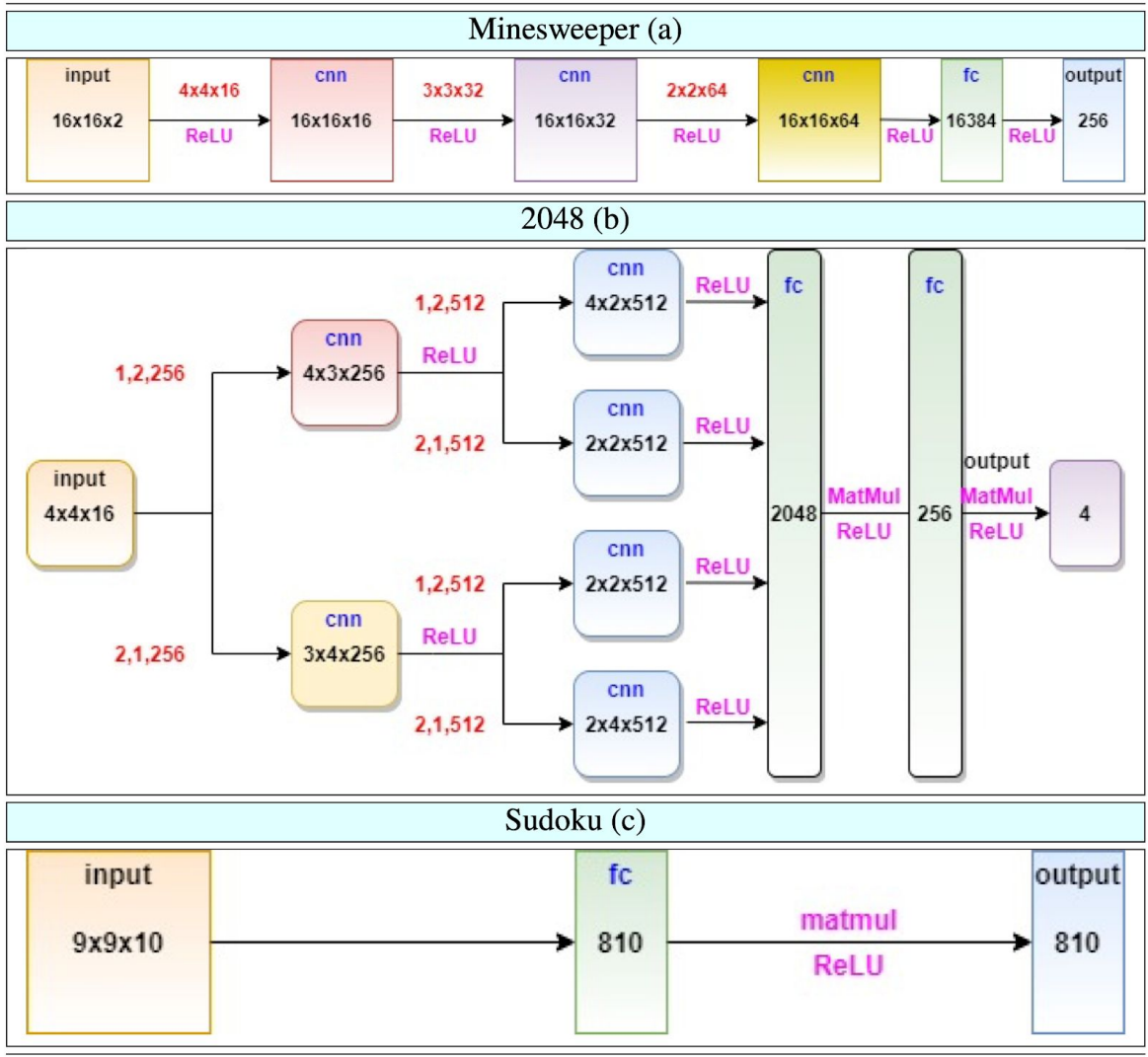

## 4. Experiments And Results

We present results for each of the games. Loss rates, rewards rates, and win rates are computed over the episodes

**15-Puzzle:** Adaptive state representation was used. For low shuffle games, the loss converges and achieves a 100% win rate. The medium and high shuffle games achieve about 43% and 22% win rates, respectively (Figure 6 (a)). There is a large jitter in rewards, which is most likely due to the reward structure, which we will address in future work.

**Minesweeper:** Average reward and win-rate for model evaluation during training and the training loss for the q-learning agent. With experience replay, it becomes possible to break any temporal correlations by mixing more and less recent experiences for the updates. The low and high density have similar profiles in loss, rewards, and win rate. Both can get to solving close to 45% of the Minesweeper. The medium density boards have a very low win rate of 15% (Figure 6(b)) because medium density boards do not open the board quickly as low-density boards, nor do they provide enough constraints as in high-density boards. Hence, these boards use guess (exploration) as much as learning.

**2048:** Training loss converged quickly in 100,000 episodes to under 0.005 (Figure 6(c)) and the 1024 win rate was achieved with significant ease (100%) with 50,000 epochs and high win rates for 2048 (40%), 4096 (0.05%), 8192 (0.01%) and 16384 (0.004%) were also achieved (Table 2).

**Sudoku:** Of the four games, Sudoku has the most challenging time in Q-learning. The loss remains high for hard and medium games and doesn't converge well. The easiest of Sudoku games had a win rate of 7%, while medium and hard games had 2.1% and 1.2% win rates, respectively (Figure 6(d)). Sudoku is an exact cover problem for boards, and hence for episodes that require multiple guesses, it is a challenging problem. Sudoku also has a large reward jitter due to the reward structure, which we will address in future research.

**Table 2: 2048 High Outcomes (Max Score and % Success)**

| 2 | 100% | 4 | 100% | 8 | 100% | 16 | 100% | 32 | 100% |
|---|---|---|---|---|---|---|---|---|---|
| 64 | 100% | 128 | 100% | 256 | 100% | 512 | 100% | 1024 | 100% |
| 2048 | 40% | 4096 | 0.05% | 8192 | 0.01% | 16384 | 0.004% | | |

# Figure 6: Results

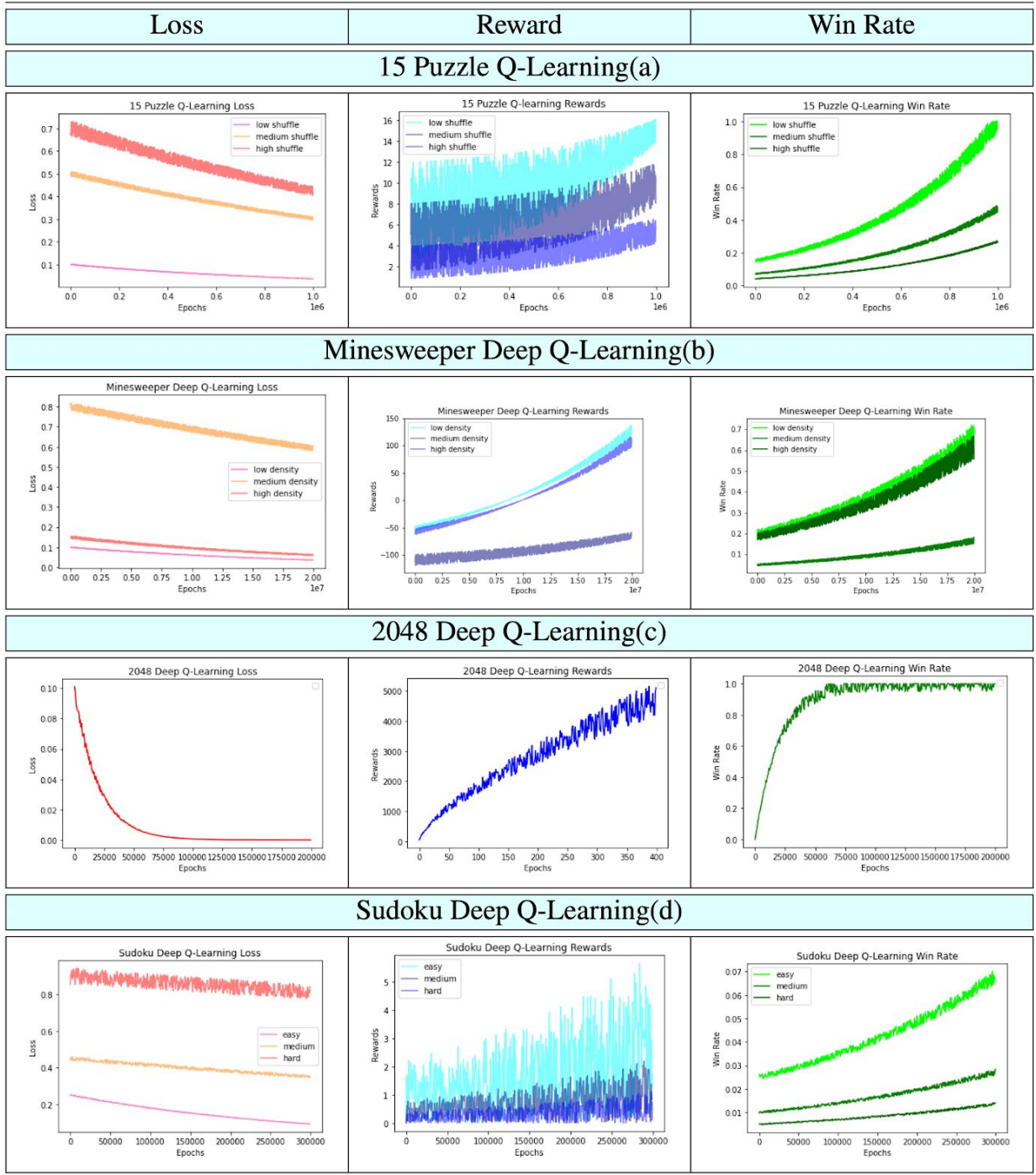

## 5. Discussion

The goal of reinforcement learning is to train an agent with reward-driven behavior on its actions. The most straight-forward reward function would be a win or lose, but this is too sparse to learn for an agent to understand its future reward. Here are a few things that could be investigated for each of the games:

**15-Puzzle:** Use a better representation of the state rather than a permutation and the reward structure to prevent jitter. Secondly, Q-learning tables from smaller puzzles (e.g., 3-Puzzle) can be scaled to bigger ones (e.g., 15-puzzle or 24-puzzle) because strategies learned from smaller problems are intuitively scalable to larger ones. And finally, use DQN implementation to filter large states.

**Minesweeper:** Q-tables from smaller minefields q-tables (e.g., 3x3) can be scaled to larger ones (e.g., 8x8), and reward structure can be changed to speed up **no_progress**, as mines free up (as clicking on space slows the convergence). A minefield can be viewed as a continuous space; hence policy gradients can be used.

**2048:** Tune rewards for better convergence of **no progress** and free space maximization. – Investigate a fully connected network (of input space (4x4x16)) instead of a CNN. Use double deep Q-Networks to mitigate overoptimistic value estimates [14] and/or dueling deep Q-network, which explicitly separates the representation of state values and state-dependent action advantages via two separate streams. [18] – Temporal Difference Learning [23, 24] is a promising area instead of a DQN. DQN can be redesigned to incorporate rotational and reflectional symmetry.

**Sudoku:** Action and rewards can be fine-tuned to encourage the reattempt of incorrectly guessed cells. Instead of using one large, fully connected network, we can investigate nine smaller grid

CNNs and one fully connected overall grid CNN. Finally, the Kaggle dataset for the Sudoku can be used for the experience replay buffer.

## 6. Conclusions

In this paper, two reinforcement learning methods, Q-learning and DQL, have been implemented and trained on 15-Puzzle, Minesweeper, 2048, and Sudoku games. Our unique contribution is in choosing the reward structure, state representation, and formulation of the DQN. Except for Sudoku, medium-density Minesweeper, and high shuffle 15-Puzzle, we successfully achieved high win rates. Since extensive networks were used, the loss is noisy, hyper-parameters were determined heuristically. It would be possible to parallelize the hyper-parameter grid search with more computational resources and yield faster convergence in training.

One of the promises of RL is that through simple learning rules such as rewards one can master complex games like Go without bias of the coder. This paper frames a set of constraint games as reinforcement learning problems and explores their environment complexity and gets us closer to understanding the limits and potential of RL and ultimately how humans learn.

## Acknowledgments

I would like to thank my parents, Dr. Reena Mehta and Dr. Huzefa Mehta, for all resources, including but not limited to Macbook Air, Google Colab Pro, and Machine Learning Coursera certificates, to pursue this project. The reinforcement learning course sparked my enthusiasm for this project. I would like to thank my brother Amal for always being my game partner. The code for this reinforcement learning project is at https://github.com/anavmehta.